\documentclass[letterpaper]{article} 
\usepackage[draft]{aaai25}  
\usepackage{times}  
\usepackage{helvet}  
\usepackage{courier}  
\usepackage[hyphens]{url}  
\usepackage{graphicx} 
\usepackage{natbib}  
\usepackage{caption} 
\usepackage{soul}

\usepackage{algorithm}
\usepackage{algorithmic}

\usepackage{newfloat}
\usepackage{listings}
\DeclareCaptionStyle{ruled}{labelfont=normalfont,labelsep=colon,strut=off} 
\floatstyle{ruled}
\newfloat{listing}{tb}{lst}{}
\floatname{listing}{Listing}
\usepackage[utf8]{inputenc} 
\usepackage{booktabs}       
\usepackage{amsfonts}       
\usepackage{nicefrac}       
\usepackage{microtype}      
\usepackage{subcaption}
\usepackage{amsmath,bm,amsthm}
\usepackage{amssymb}
\usepackage{makecell}
\usepackage{xcolor}
\usepackage[export]{adjustbox}
\usepackage{array}
\newcolumntype{H}{>{\setbox0=\hbox\bgroup}c<{\egroup}@{}}
\usepackage{multirow}

\DeclareCaptionFont{tenpt}{\fontsize{10pt}{12pt}\selectfont #1}
\definecolor{darkpastelgreen}{rgb}{0.01, 0.75, 0.24}
\definecolor{darkorange}{rgb}{1.0, 0.55, 0.0}

\newcommand{\RE}[1]{\textcolor{black}{#1}}

\newcommand{\E}{\mathbb{E}}
\newcommand{\R}{\mathbb{R}}

\newcommand{\boldx}{\boldsymbol{x}}

\newcommand{\calD}{\mathcal{D}}

\newcommand{\calL}{\mathcal{L}}

\newcommand{\calN}{\mathcal{N}}

\newcommand{\indicator}{\mathbf{1}}

\newcommand{\bgamma}{\boldsymbol{\gamma}}

\newcommand{\bsigma}{\boldsymbol{\sigma}}
\newcommand{\bphi}{\boldsymbol{\phi}}

\newcommand{\btheta}{\boldsymbol{\theta}}
\newcommand{\bmu}{\boldsymbol{\mu}}

\title{Sequential Bayesian Neural Subnetwork Ensembles}
\author{
    Sanket Jantre\textsuperscript{\rm 1}\thanks{Corresponding authors}, Shrijita Bhattacharya\textsuperscript{\rm 2}, Nathan M. Urban\textsuperscript{\rm 1},\\ Byung-Jun Yoon\textsuperscript{\rm 1,3}, Tapabrata Maiti\textsuperscript{\rm 2}, Prasanna Balaprakash\textsuperscript{\rm 4}, Sandeep Madireddy\textsuperscript{\rm 5 *}
}
\affiliations{
    \textsuperscript{\rm 1} Computational Science Initiative, Brookhaven National Laboratory \\
    \textsuperscript{\rm 2} Department of Statistics and Probability, Michigan State University \\
    \textsuperscript{\rm 3} Department of Electrical and Computer Engineering, Texas A\&M University \\
    \textsuperscript{\rm 4} Computing and Computational Sciences Directorate, Oak Ridge National Laboratory \\
    \textsuperscript{\rm 5} Mathematics and Computer Science Division, Argonne National laboratory\\
    \{sjantre, nurban, byoon\}@bnl.gov, \{bhatta61, maiti\}@msu.edu, pbalapra@ornl.gov, smadireddy@anl.gov
}

\begin{document}

\maketitle

\begin{abstract}
Deep ensembles have emerged as a powerful technique for improving predictive performance and enhancing model robustness across various applications by leveraging model diversity. However, traditional deep ensemble methods are often computationally expensive and rely on deterministic models, which may limit their flexibility. Additionally, while sparse subnetworks of dense models have shown promise in matching the performance of their dense counterparts and even enhancing robustness, existing methods for inducing sparsity typically incur training costs comparable to those of training a single dense model, as they either gradually prune the network during training or apply thresholding post-training. In light of these challenges, we propose an approach for sequential ensembling of dynamic Bayesian neural subnetworks that consistently maintains reduced model complexity throughout the training process while generating diverse ensembles in a single forward pass. Our approach involves an initial exploration phase to identify high-performing regions within the parameter space, followed by multiple exploitation phases that take advantage of the compactness of the sparse model. These exploitation phases quickly converge to different minima in the energy landscape, corresponding to high-performing subnetworks that together form a diverse and robust ensemble. We empirically demonstrate that our proposed approach outperforms traditional dense and sparse deterministic and Bayesian ensemble models in terms of prediction accuracy, uncertainty estimation, out-of-distribution detection, and adversarial robustness.
\end{abstract}

%


\section{Introduction} 
\label{sec:intro}
Deep learning has powere state-of-the-art performance in a wide array of machine learning tasks \cite{LeCun-et-al-2015}. However, deep learning models still face many fundamental issues from the perspective of statistical modeling, which are crucial in many fields including autonomous driving, healthcare, and science \cite{bartlett-2021}. One of the major challenges is their ability to reliably estimate model uncertainty while capturing complex data dependencies and being computationally tractable. Probabilistic machine learning, especially the Bayesian framework, offers an exciting avenue to address these challenges. Besides superior uncertainty quantification, Bayesian models exhibit improved robustness to noise and adversarial perturbations \cite{wicker2021bayesian} due to probabilistic prediction capabilities. Bayesian neural networks (BNNs) have pushed the envelope of probabilistic machine learning through the combination of deep neural network (DNN) architecture and Bayesian inference. However, due to the enormous number of parameters, BNNs adopt approximate inference techniques such as variational inference with a fully factorized approximating family \cite{Jordan_Graph-2000}. Although this approximation is crucial for computational tractability, it may under-utilize BNN's true potential \cite{Izmailov-et-al-2021}.

\begin{figure}
    \centering
    \includegraphics[width=0.98\linewidth]{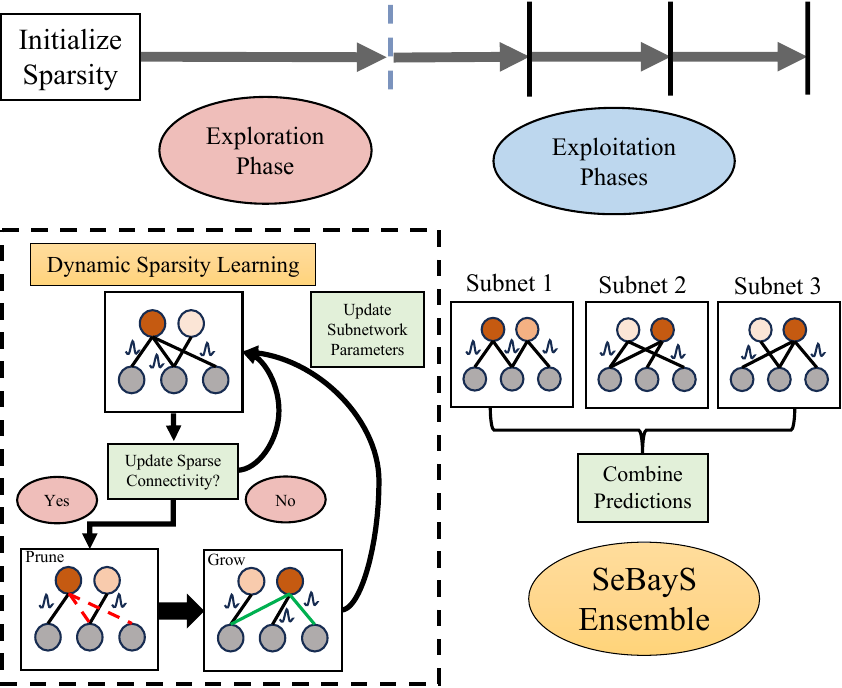}
    \caption{Illustration of {\tt SeBayS} ensemble: Our approach includes an {\it exploration phase} followed by multiple {\it exploitation phases} to create Bayesian subnetworks. {\tt SeBayS} ensemble prediction is obtained by combining their predictions.}
    \label{fig:SeBayS_illustration}
\end{figure}

Ensembles of neural networks \cite{lakshminarayanan2017simple} have been proposed to account for the parameter/model uncertainty, which are shown to be analogous to Bayesian model averaging and sampling from the parameter posteriors to estimate the posterior predictive distribution \cite{wilson2020bayesian}. In this spirit, ensemble diversity is a key to enhancing predictions, uncertainty, and robustness. To this end, diverse ensembles can mitigate shortcomings of approximate Bayesian inference without sacrificing computational efficiency. In past, various diversity-inducing techniques have been explored, including using specific learning rate schedule \cite{huang2017snapshot}, kernalized repulsion terms in the loss function \cite{d2021repulsive}, mixtures of approximate posteriors capturing multiple posterior modes \cite{RANK-1-BNN}, sparsity as a mechanism for diversity~\cite{MIMO,FREE-TICKETS}, and diversity in model architectures via neural architecture and hyperparameter searches \cite{egele2021autodeuq,wenzel2020hyperparameter}.

However, most approaches use parallel ensembles, where individual model part starts with a different initialization. This can be computationally expensive, as each ensemble member requires extended training to reach a high-performing neighborhood of the parameter space. Although ensemble diversity is emphasized, the training cost is often overlooked. With the size of models only growing as we advance in deep learning, reducing the training cost of ensemble models alongside increasing their diversity is crucial.

Sequential ensembling offers an elegant solution to reduce the cost of obtaining multiple ensembles, with roots in methods that combine learning trajectory epochs \cite{swann1998fast, xie2013horizontal}. \cite{jean2014using, sennrich2016edinburgh} leverage intermediate training stages, while \cite{moghimi2016boosted} uses boosting to generate ensembles. Recent methods~\cite{huang2017snapshot,FGE,FREE-TICKETS} employ cyclic learning rate annealing to force the model to visit multiple local minimas, collecting ensembles at each local minimum. All these techniques have been primarily applied to deterministic models. Extending sequential ensembling to Bayesian models is attractive as it can allow creation of high-performing ensembles without the need to train from scratch, similar to sampling a posterior distribution with a Markov chain Monte Carlo sampler. Sequential ensembling can also complement parallel ensembling, where each parallel ensemble can generate multiple sequential ensembles, increasing the overall diversity of the final ensemble model.

A new frontier for improving the computational tractability and robustness of neural networks is {\it sparsity}~\cite{hoefler2021sparsity}. Famously, the lottery ticket hypothesis~\cite{frankle2018LOT} established the existence of sparse subnetworks that can match the performance of the dense model. Studies have also shown that such subnetworks tend to be inherently diverse due to different neural connectivity~\cite{MIMO,FREE-TICKETS,yin2023lottery}. However, most sparsity-inducing techniques have focused on deterministic networks, using post-hoc pruning \citep{Han-et-al-2016,Molchanov-et-al-2017}. In Bayesian learning, the prior distribution provides a systematic approach to incorporate inductive bias and expert knowledge directly into the model~\cite{robert2007bayesian}. Consequently, sparsity in Bayesian neural networks can be introduced via sparsity-inducing priors~\cite{Louizos-et-al-2017,Bai-Guang-2020,Ghosh-JMLR-2018, jantre2023shrinkage, jantre2023layer} which either incorporate sparsity gradually during training or post-training using thresholding criteria. These approaches successfully reduce computational and memory costs during inference, but their training costs remain similar to those of dense models. 

To this end, we propose {\bf Se}quential {\bf Bay}esian Neural {\bf S}ubnetwork Ensembles ({\tt SeBayS}) with the following key contributions:
\begin{itemize}
    \item We propose a sequential ensembling strategy for Bayesian neural networks (BNNs) which learns multiple subnetworks in a single forward-pass utilizing a fully sparse Bayesian framework that embeds sparsity in the posterior from the start of training. The approach involves a single exploration phase to find high-performing sparse network connectivity followed by multiple exploitation phases to obtain multiple subnetworks for ensembling.
    \item We leverage the light-weight dynamic sparsity learning to efficiently generate diverse sparse Bayesian neural networks, which we refer to as {\it Bayesian neural subnetworks}. Our approach outperforms current state-of-the-art methods in terms of predictive accuracy, uncertainty estimation, out-of-distribution detection, and adversarial robustness.
\end{itemize}


\section{Related Work} 
\label{sec:rel-work}
\textbf{Ensembles of neural networks:}
Ensembling techniques in the context of neural networks are increasingly being adopted in the literature due to their potential to improve accuracy, robustness, and quantify uncertainty. The most simple and widely used approach is Monte Carlo dropout, which is based on Bernoulli noise~\cite{gal2016dropout} and deactivates certain units during training and testing. This, along with techniques such as DropConnect~\cite{wan2013regularization}, Swapout~\cite{singh2016swapout} are referred to as``implicit" ensembles as model ensembling is happening internally in a single model. Although they are efficient, the gain in accuracy and robustness is limited, and they are mainly used in the context of deterministic models. Although most recent approaches have targeted parallel ensembling techniques, few approaches such as BatchEnsemble~\cite{Wen2020BatchEnsemble} appealed to parameter efficiency by decomposing ensemble members into a product of a shared matrix and a rank-one matrix, and using the latter for ensembling and MIMO~\cite{MIMO} which discovers subnetworks from a larger network via multi-input multi-output configuration. In the context of Bayesian neural network ensembles,~\cite{RANK-1-BNN} proposed a rank-1 parameterization of BNNs, where each weight matrix involves only a distribution on a rank-1 subspace and uses mixture approximate posteriors to capture multiple modes. \cite{premchandarunified} employ weight space ensembling with distribution learning of architectures for improved robustness.

Sequential ensembling techniques offer an elegant solution to ensemble training but have not received much attention recently due to a wider focus of the community on diversity of ensembles and less on the computational cost. Notable sequential ensembling techniques are ~\cite{huang2017snapshot,FGE,FREE-TICKETS} that enable the model to visit multiple local minima through cyclic learning rate annealing and collect ensembles only when the model reaches a local minimum. The difference is that ~\cite{huang2017snapshot} adopts cyclic cosine annealing, ~\cite{FGE} uses a piece-wise linear cyclic learning rate schedule that is inspired by geometric insights. Finally, ~\cite{FREE-TICKETS} adopts a piece-wise constant cyclic learning rate schedule. We also note that all of these approaches have been primarily in the context of deterministic neural networks. 

Our approach (i) introduces sequential ensembling into Bayesian neural networks, (ii) combines it with dynamic sparsity learning for cheaply collecting Bayesian subnetworks, and (iii) efficiently produces diverse model ensembles. It complements other parallel and efficient ensemble methods.

\section{Sequential Bayesian Neural Subnetwork Ensembles}
\label{sec:methods}

\subsection{Preliminaries}
\vspace{0.5mm}
\textbf{Bayesian Neural Networks.} Let $\mathcal{D}=\{(\boldx_i,y_i)\}_{i=1,\cdots,N}$ represent a training dataset of $N$ i.i.d. observations, where $\boldx$ denotes input samples and $y$ denotes corresponding outputs. In the Bayesian framework, instead of optimizing over a single probabilistic model, $p(y|\boldx,\btheta)$, we discover all likely models via posterior inference over model parameters, $\btheta = (\theta_1,\cdots,\theta_T) \in \R^T$. The Bayes' rule provides the posterior distribution: $p(\btheta|\mathcal{D}) \propto p(\mathcal{D}|\btheta) p(\btheta)$, where $p(\mathcal{D}|\btheta)$ denotes the likelihood of $\mathcal{D}$ given $\btheta$ and $p(\btheta)$ is the prior over parameters. Using $p(\btheta|\mathcal{D})$, we predict the label for a new example $\boldx^*$
through Bayesian model averaging:
\begin{align*}
p(y^*|\boldx^*,\calD) &= \int p(y^*|\boldx^*,\btheta) p(\btheta|\calD) d \btheta \\
& \approx \frac{1}{M} \sum_{m=1}^M p(y^*|\boldx^*,\btheta_m) , \enskip \enskip \btheta_m \sim p(\btheta|\calD)
\end{align*}

\vspace{1mm}
\noindent \textbf{Variational Inference (VI).} Although, Markov chain Monte Carlo sampling is the gold standard for Bayesian inference, it is computationally inefficient \cite{Izmailov-et-al-2021}. Variational inference, in contrast, is faster and scales well for complex tasks with large datasets \cite{Blei2017}. Variational learning infers a distribution $q(\btheta)$ on model parameters that minimises the Kullback-Leibler (KL) distance from the true posterior $p(\btheta|\mathcal{D})$:
$$ \widehat{q}(\btheta)=\underset{q(\btheta) \in \mathcal{Q}}{\text{argmin}}\:\: d_{\rm KL}(q(\btheta),p(\btheta|\mathcal{D}))$$
where $\mathcal{Q}$ denotes the variational family of distributions. This optimization problem is equivalent to minimizing the negative Evidence Lower Bound (ELBO), defined as
\begin{equation}
\label{e:elbo} 
    \mathcal{L}= -\mathbb{E}_{q(\btheta)} [\log p(\mathcal{D}|\btheta)]+d_{\rm KL}(q(\btheta),p(\btheta)),
\end{equation}
where the first term is the data-dependent cost, known as the negative log-likelihood (NLL), and the second term is prior-dependent and serves as regularization. The NLL is often intractable and estimated using Monte Carlo sampling. Direct optimization of \eqref{e:elbo} is computationally prohibitive, so gradient descent methods are employed \cite{Kingma-welling-2014}, with the reparameterization trick used for efficient gradient backpropagation.

\vspace{1.5mm}
\noindent \textbf{Prior Choice.}
Dynamic sparsity learning to obtain Bayesian subnetworks is achieved by randomly selecting a sparse connectivity of preferred dimension $s$. Thus, a prior on $\btheta$ that incorporates this sparsity can be defined as follows:
\begin{equation*}
    p(\theta_i) = \gamma_i \calN(0,\sigma_0^2), \enskip \gamma_i = \indicator_S(i) \hspace{1mm} {\rm with} \hspace{1mm}  S = \{n_1,\cdots,n_s\}
\end{equation*}
Here, $\indicator_S(.)$ is the indicator function on the set $S$ of non-zero weight indices. The variational family mirrors the prior's structure to maintain sparsity in variational approximation. The mean-field variational family structure is:
\begin{equation*}
    q_{\phi_i}(\theta_i) = \gamma_i \calN(\mu_i,\sigma_i^2), \enskip \gamma_i = \indicator_S(i),  \hspace{1mm}  S = \{n_1,\cdots,n_s\}
\end{equation*}
here, $\phi_i=(\mu_i,\sigma_i^2)$ represents the variational mean and variance parameters of $q_{\phi_i}(\theta_i)$. For sparse learning, $(\mu_i,\sigma_i^2) = (0,0)$ when $i \notin \{n_1,\cdots,n_s\}$. We derive the ELBO from (\ref{e:elbo}) to formulate the optimization problem as minimizing
\begin{equation}
\label{e:loss}
    \calL = -\mathbb{E}_{q_{\bphi}(\btheta)} [\log p(\mathcal{D}|\btheta)] + \sum_{i=1}^T \gamma_i d_{\rm KL}(q_{\phi_i}(\theta_i),p(\theta_i))
\end{equation}
This formulation maintains the target sparsity level $s$ throughout model training, reducing computational and memory demands. Periodically, we explore the parameter space to enhance sparse connectivity by pruning a fraction $p$ of the least significant weights from $\btheta_s$, followed by regrowing an equivalent number of weights during training. This approach improves upon static sparsity methods while preserving the same computational budget.

\subsection{Bayesian Neural Subnetworks}
We follow the key steps of the deterministic dynamic sparsity learning method, Rigged Lottery \cite{evci2020rigging}: initial sparse connectivity, model weight optimization, and the prune-grow criterion to collect Bayesian subnetworks.

\vspace{1.5mm}
\noindent {\bf Initial Sparse Connectivity.} We initialize the sparsity with {\it Erd{\H o}s-R{\'e}nyi-Kernel} (ERK) method, which scales the sparsity of the $l^{\rm th}$-convolution layer by a factor of $1-\frac{n^{l-1}+n^l+w^l+h^l}{n^{l-1} \times n^l \times w^l \times h^l}$, where $w^l$ and $h^l$ are the width and height of the $l^{\rm th}$-layer convolution kernel and $n^l$ are the number of channels in the $l^{\rm th}$- convolution layer. For the $l^{\rm th}$-linear layer, the sparsity scales with a factor of $1-\frac{n^{l-1}+n^l}{n^{l-1} \times n^l}$. This sparse allocation ensures that  layers with more parameters are pruned more aggressively. Once the initial set of sparse indices $S_{\rm init}$ is selected, we initialize the $\bphi_0$ on $S_{\rm init}$ following the procedure used in VI-based BNN techniques. 

\vspace{1.5mm}
\noindent {\bf Model Weight Optimization.} 
For a given $\bgamma$, $\bphi$ is optimised using the standard VI-based BNN training approach with the SGD algorithm for $\Delta T$ steps as outlined in \cite{li2024ssvi}. During the forward pass through the sparse BNN, the NLL term in (\ref{e:loss}) is estimated via Monte Carlo sampling from $q_{\bphi}(\btheta)$, employing local-reparameterization trick for efficiency \cite{kingma2015lrt}. After $\Delta T$ steps, Lastly, upon updating the sparse connectivity ($\bgamma^\tau \to \bgamma^{\tau+1}$), the newly included non-zero weights (denoted by indices $j$) initially have zero $\sigma_j^{\tau+1}$ values that impact the gradient descent since their gradients are also zero. We set $\sigma_j^{\tau+1}$s to the average of the non-zero $\sigma_k^\tau$ values from the same convolution kernel or fully connected layer as $\sigma_j^{\tau+1}$.

\vspace{1.5mm}
\noindent {\bf Prune-Grow Criterion.} 
We update $\bphi$ and $\bgamma$ using an alternating optimization strategy. After every $\Delta T$ steps, the sparse connectivity $\bgamma$ is updated deterministically as described in \cite{li2024ssvi}. Specifically, $\bgamma^{\tau+1}$ is obtained by pruning a fraction $p$ of less significant weight indices from $\bgamma^\tau$ based on $\bphi_\tau = (\bmu^\tau,{\bsigma^\tau}^2)$. In particular, we compute the Signal-to-Noise (SNR) of $|\btheta|$ to consider both the magnitude and variance of the weights during pruning, similar to \cite{kingma2015lrt}. Additional details on SNR($|\btheta|$) are provided in the Appendix~\ref{App:prune-grow}. 

After pruning step, we reintroduce the same fraction of weight indices into the sparse connectivity $\bgamma^{\tau+1}$ for the next $\Delta T$ steps. The method uses the highest absolute gradients of the weights, computed with a batch of inputs $\boldx$ of size $B$ and stochastic weight sample $\btheta$. The grow criterion is given by:
$$ \E_{q_{\bphi}} \left | \E_{\boldx}\frac{1}{B} \sum_{i=1}^{B} \nabla_{\btheta} \calL (x_i) \right |.  $$
We perform a one-step MC estimation to approximate the double expectation above, resulting in $|  \sum_{i=1}^{B} \nabla_{\btheta} \calL (x_i)|/B$.

We chose $\Delta T=1000$ in our experiments to be consistent with \cite{FREE-TICKETS,li2024ssvi}.

\subsection{Sequential Ensembling Strategy}
We propose a sequential ensembling procedure to obtain base learners (individual models part of an ensemble) $\{\btheta^1,\btheta^2,\cdots,\btheta^M\}$ that are collected in a single training run and used to construct the ensemble. The ensemble predictions are computed by averaging the predictions from each base learner. Specifically, if $y^m_{\rm new}$ denotes the outcome from the $m^{\rm th}$ base learner, then the ensemble prediction for $M$ base learners (for continuous outcomes) is given by $y_{\rm new} = \frac{1}{M} \sum_{m=1}^M y^m_{\rm new}$.

Our ensembling strategy generates a diverse set of base learners through a single end-to-end training process, which includes an exploration phase followed by $M$ exploitation phases. During the exploration phase, we use a large constant learning rate for $T_0$ time steps to explore high-performing regions of the parameter space, aiming for sparse connectivity with strong predictive performance. At the conclusion of the exploration phase, the model sparsity and corresponding variational posterior parameters reach a good region on the posterior density surface.

Next, in each equally spaced exploitation phase $({\rm time}=t_{\rm ex})$ of the ensemble training, we follow a two-step learning rate schedule: first applying a moderately large learning rate for $t_{\rm ex}/2$ time followed by a small learning rate for the remaining $t_{\rm ex}/2$ time. After the first model converges $({\rm time}=t_0+t_{\rm ex})$, which gives us our first sparse base learner, we prune a large fraction $p_L$ of the sparse connectivity and then regrow same fraction of the weight indices according to our grow criterion. This approach helps the model escape current local minima and find a new performant subnetwork. We repeat this process $M-1$ times to generate $M$ sparse base learners, including the first learner, which does not undergo the large prune-grow phase. Combining these individual sparse base learner results in our Sequential Bayesian Neural Subnetwork Ensemble (\texttt{SeBayS}). The pseudocode for {\tt SeBayS} ensemble is provided in Algorithm~\ref{alg_Seq_Ens}.

\begin{algorithm}[t!]
\caption{{\bf Se}quential {\bf Bay}esian Neural {\bf S}ubnetwork Ensembles (\texttt{SeBayS})} 
\label{alg_Seq_Ens}
\begin{algorithmic}[1]
    \STATE {\bfseries Inputs:} training data $\mathcal{D}=\{(\boldx_i,y_i)\}_{i=1}^N$, ensemble size $M$, target sparsity $s$, set of initial sparse indices $S_{\rm init}$, update interval $\Delta T$, prune-grow rate $p$, large prune-grow rate $p_L$, exploration phase training time $t_0$, training time of each exploitation phase $t_{\rm ex}$. \\
    \hspace{1mm} \textit{Model inputs:} prior hyperparameter $\sigma_0^2$.
    \STATE {\bfseries Output:} Variational parameters $\bphi= (\bmu,\bsigma^2)$ estimates and sparse connectivities $\bgamma$.
    \STATE {\bfseries Method:} Set initial sparse connectivity $\gamma_{\rm init}$ using $S_{\rm init}$; initialize variational parameters: $\bphi_{\rm init}=(\bmu_{\rm init}, \bsigma_{\rm init}^2)$. \\
    \textcolor{gray}{\# Exploration Phase}
    \FOR{$t= 1,2,\dots, T_0 $}
        \STATE Update $\bphi^0$ $\leftarrow$ SGD($\mathcal{L}$).
        \IF{($t$ mod $\Delta T$) = 0}
            \STATE Update $\bgamma^0$ using prune-grow rate $p$.
        \ENDIF
    \ENDFOR
    \\\textcolor{gray}{\# M Sequential Exploitation Phases}
    \FOR{$m= 1,2,\dots, M$}
        \FOR{$t= 1,2,\dots, t_{\rm ex}$}
            \STATE Update $\bphi^m$ $\leftarrow$ SGD($\mathcal{L}$).
            \IF{($t$ mod $\Delta T$) = 0}
                \STATE Update $\bgamma^m$ using prune-grow rate $p$.
            \ENDIF
        \ENDFOR
        \STATE Save the model with weights $\btheta^m$ sampled using $\bphi^m$.
        \STATE Update $\bgamma^m \to \bgamma^{m+1}$ using large prune-grow rate $p_L$.
    \ENDFOR
\end{algorithmic}
\end{algorithm}

We also perform ensembling over Bayesian subnetworks obtained in parallel, where each subnetwork undergoes a single exploration phase followed by a single exploitation phase. This results in a parallel Bayesian Neural Subnetwork Ensemble, which we refer to as the {\tt BayS} ensemble. For baseline comparison, we also include a parallel ensemble of dense BNN models trained using the same strategy as the {\tt BayS} ensemble, but with dense connectivity maintained throughout. We refer to this as the dense BNN ensemble. 

\section{Experimental Results}
\label{sec:CIFAR_exp}


\begin{table*}[ht]
\fontsize{9}{11}\selectfont
\centering
\begin{tabular}{l*{6}{c}*{1}{H}*{1}{c}}
\toprule
Methods & Acc ($\uparrow$) & NLL ($\downarrow$) & ECE ($\downarrow$) & cAcc ($\uparrow$) & cNLL ($\downarrow$) & cECE ($\downarrow$) & \makecell{\# Training \\ FLOPs ($\downarrow$)} ($\downarrow$) & \makecell{\# Training \\ runs ($\downarrow$)} \\
\midrule
Single Dense DNN Model* & 96.0 & 0.159 & 0.023 & 76.1 & 1.050 & 0.153 & 3.6e17 & 1 \\
\RE{Single Dense BNN Model} & \RE{95.8} & \RE{0.170} & \RE{0.024} & \RE{75.8} & \RE{1.146} & \RE{0.161} & -- & 1 \\
\RE{SeBayS (M = 1) (S = 0.8)} & \RE{95.8} & \RE{0.162} & \RE{0.022} & \RE{76.6} & \RE{0.999} & \RE{0.141} & -- & 1 \\
\RE{SeBayS (M = 1) (S = 0.9)} & \RE{95.4} & \RE{0.158} & \RE{0.023} & \RE{76.8} & \RE{0.953} & \RE{0.137} & -- & 1 \\
\RE{BayS (M = 1) (S = 0.8)} & \RE{95.6} & \RE{0.170} & \RE{0.025} & \RE{76.2} & \RE{1.123} & \RE{0.157} & -- & 1 \\
Monte Carlo Dropout* & 95.9 & 0.160 & 0.024 & 68.8 & 1.270 & 0.166 & 1.00× & 1 \\
MIMO (M = 3)* & \textbf{96.4} & 0.123 & 0.010 & 76.6 & 0.927 & 0.112 & 1.00× & 1 \\
EDST Ensemble (M = 3) (S = 0.8)* & 96.3 & 0.127 & 0.012 & 77.9 & 0.814 & 0.093 & 0.61× & 1 \\
EDST Ensemble (M = 7) (S = 0.9)* & 96.1 & 0.122 & 0.008 & 77.2 & 0.803 & 0.081 & 0.57× & 1 \\
\RE{SeBayS Ensemble (M = 3) (S = 0.8)} & \RE{96.2} & \RE{0.129} & \RE{0.009} & \RE{77.7} & \RE{0.842} & \RE{0.092} & -- & 1 \\
\RE{SeBayS Ensemble (M = 7) (S = 0.9)} & \RE{96.2} & \RE{\textbf{0.118}} & \RE{\textbf{0.006}} & \RE{\textbf{78.0}} & \RE{\textbf{0.776}} & \RE{\textbf{0.080}} & -- & 1 \\
\midrule
TreeNet (M = 3)* & 95.9 & 0.158 & 0.018 & 75.6 & 0.969 & 0.137 & 1.52× & 1.5 \\
BatchEnsemble (M = 4)* & 96.2 & 0.143 & 0.021 & 77.5 & 1.020 & 0.129 & 1.10× & 4 \\
Rank-1 BNN (M = 4)$^\dagger$ & 96.3 & 0.128 & \textcolor{blue}{\textbf{0.008}} & 76.7 & 0.840 & 0.080 & -- & 4 \\
LTR Ensemble (M = 3) (S = 0.8)* & 96.2 & 0.133 & 0.015 & 76.7 & 0.950 & 0.118 & 1.75× & 4 \\
Static Sparse Ensemble (M = 3) (S = 0.8)* & 96.0 & 0.133 & 0.014 & 76.2 & 0.920 & 0.098 & \textbf{1.01×} & 3 \\
PF Ensemble (M = 3) (S = 0.8)* & 96.4 & 0.129 & 0.011 & 78.2 & 0.801 & 0.082 & 3.75× & 6 \\
DST Ensemble (M = 3) (S = 0.8)* & 96.3 & \textcolor{blue}{\textbf{0.122}} & 0.010 & \textcolor{blue}{\textbf{78.8}} & \textcolor{blue}{\textbf{0.766}} & \textcolor{blue}{\textbf{0.075}} & \textcolor{blue}{\textbf{1.01×}} & 3 \\
\RE{BayS Ensemble (M = 3) (S = 0.8)} & \RE{\textcolor{blue}{\textbf{96.5}}} & \RE{0.127} & \RE{0.010} & \RE{78.2} & \RE{0.852} & \RE{0.095} & -- & 3 \\
\midrule
Dense DNN Ensemble (M = 4)* & 96.6 & 0.114 & 0.010 & 77.9 & 0.810 & 0.087 & 4.00× & 4 \\
\RE{Dense BNN Ensemble (M = 3)} & \RE{96.5} & \RE{0.125} & \RE{0.010} & \RE{78.1} & \RE{0.872} & \RE{0.095} & -- & 3 \\
\bottomrule
\end{tabular}
\caption{Wide ResNet28-10/CIFAR-10: we mark the best results of one-pass efficient ensemble in bold and multi-pass efficient ensemble in blue. Results with * are obtained from \cite{FREE-TICKETS}. $^\dagger$ Rank-1 BNN results are from \cite{RANK-1-BNN}.}
\label{table:wide_resnet_cifar_10}
\end{table*}


\begin{table*}[ht]
\fontsize{9}{11}\selectfont
\centering
\begin{tabular}{l*{6}{c}*{1}{H}*{1}{c}}
\toprule
Methods & Acc ($\uparrow$) & NLL ($\downarrow$) & ECE ($\downarrow$) & cAcc ($\uparrow$) & cNLL ($\downarrow$) & cECE ($\downarrow$) & \makecell{\# Training \\ FLOPs ($\downarrow$)} ($\downarrow$) & \makecell{\# Training \\ runs ($\downarrow$)} \\
\midrule
Single Dense DNN Model* & 79.8 & 0.875 & 0.086 & 51.4 & 2.700 & 0.239 & 3.6e17 & 1 \\
\RE{Single Dense BNN Model} & \RE{80.3} & \RE{0.825} & \RE{0.066} & \RE{52.0} & \RE{2.451} & \RE{0.204} & -- & 1 \\
\RE{SeBayS (M = 1) (S = 0.8)} & \RE{80.1} & \RE{0.805} & \RE{0.060} & \RE{51.7} & \RE{2.355} & \RE{0.174} & -- & 1 \\
\RE{SeBayS (M = 1) (S = 0.9)} & \RE{79.8} & \RE{0.836} & \RE{0.058} & \RE{50.5} & \RE{2.473} & \RE{0.180} & -- & 1 \\
\RE{BayS (M = 1) (S = 0.8)} & \RE{79.7} & \RE{0.814} & \RE{0.058} & \RE{51.6} & \RE{2.443} & \RE{0.181} & -- & 1 \\
Monte Carlo Dropout* & 79.6 & 0.830 & 0.050 & 42.6 & 2.900 & 0.202 & 1.00× & 1 \\
MIMO (M = 3)* & 82.0 & 0.690 & 0.022 & 53.7 & 2.284 & 0.129 & 1.00× & 1 \\
EDST Ensemble (M = 3) (S = 0.8)* & 82.2 & 0.672 & 0.034 & \textbf{54.0} & 2.156 & 0.137 & 0.61× & 1 \\
EDST Ensemble (M = 7) (S = 0.9)* & \textbf{82.6} & \textbf{0.653} & 0.036 & 52.7 & 2.410 & 0.170 & 0.57× & 1 \\
\RE{SeBayS Ensemble (M = 3) (S = 0.8)} & \RE{81.5} & \RE{0.701} & \RE{0.024} & \RE{53.5} & \RE{\textbf{2.148}} & \RE{0.119} & -- & 1 \\
\RE{SeBayS Ensemble (M = 7) (S = 0.9)} & \RE{81.8} & \RE{0.676} & \RE{\textbf{0.019}} & \RE{53.1} & \RE{2.205} & \RE{\textbf{0.112}} & -- & 1 \\
\midrule
TreeNet (M = 3)* & 80.8 & 0.777 & 0.047 & 53.5 & 2.295 & 0.176 & 1.52× & 1.5 \\
BatchEnsemble (M = 4)* & 81.5 & 0.740 & 0.056 & 54.1 & 2.490 & 0.191 & 2.10× & 4 \\
Rank-1 BNN (M = 4)$^\dagger$ & 81.3 & 0.692 & 0.018 & 53.8 & 2.240 & 0.117 & -- & 4 \\
LTR Ensemble (M = 3) (S = 0.8)* & 82.2 & 0.703 & 0.045 & 53.2 & 2.345 & 0.180 & 1.75× & 4 \\
Static Sparse Ensemble (M = 3) (S = 0.8)* & 82.4 & 0.691 & 0.035 & 52.5 & 2.468 & 0.167 & 1.01× & 4 \\
PF Ensemble* & 83.2 & 0.639 & 0.020 & 54.2 & 2.182 & 0.115 & 3.75× & 3 \\
DST Ensemble (M = 3) (S = 0.8)* & \textcolor{blue}{\textbf{83.3}} & \textcolor{blue}{\textbf{0.623}} & 0.018 & \textcolor{blue}{\textbf{55.0}} & 2.109 & 0.104 & 1.01× & 3 \\
\RE{BayS Ensemble (M = 3) (S = 0.8)} & \RE{82.0} & \RE{0.664} & \RE{\textcolor{blue}{\textbf{0.016}}} & \RE{54.8} & \RE{\textcolor{blue}{\textbf{2.101}}} & \RE{\textcolor{blue}{\textbf{0.096}}} & -- & 3 \\
\midrule
Dense DNN Ensemble (M = 4)* & 82.7 & 0.666 & 0.021 & 54.1 & 2.270 & 0.138 & 4.00× & 4 \\
\RE{Dense BNN Ensemble (M = 3)} & \RE{82.5} & \RE{0.670} & \RE{0.021} & \RE{54.7} & \RE{2.151} & \RE{0.125} & -- & 3 \\
\bottomrule
\end{tabular}
\caption{Wide ResNet28-10/CIFAR-100: we mark the best results of one-pass efficient ensemble in bold and multi-pass efficient ensemble in blue. Results with * are obtained from \cite{FREE-TICKETS}. $^\dagger$ Rank-1 BNN results are from \cite{RANK-1-BNN}.}
\label{table:wide_resnet_cifar_100}
\end{table*}

In this section, we demonstrate the performance of our proposed \texttt{SeBayS} approach on network architectures and datasets commonly used in practice. Specifically, we use Wide ResNet28-10 model \cite{wrn} and train on the CIFAR-10 and CIFAR-100 datasets \cite{krizhevsky2009cifar}. The models are trained with batch normalization, step-wise (piece-wise constant) learning rate schedules, and augmented training data. To ensure stable training, we applied KL annealing \cite{sonderby2016ladder} and used a smaller learning rate for the variational variance parameter $\bsigma^2$ during exploration phase. We also conducted an extensive hyperparameter search, with the final settings provided in Appendix~\ref{App:Expt_details}. The details on fairness, uniformity, consistency in training and evaluation, and reproducibility considerations for \texttt{SeBayS} and other models are provided in Appendix~\ref{App:Expt_details}.

\vspace{2mm}
\noindent \textbf{Baselines.} Our baselines include a single deterministic deep neural network (DNN), variational inference trained single Bayesian neural network (BNN) \cite{blundell2015weight} (the top-performing BNN from the dense BNN ensemble), and the best individual Bayesian neural subnetworks from both the {\tt BayS} and {\tt SeBayS} ensembles. We also benchmark our method against several ensembling strategies, including Monte Carlo Dropout \cite{gal2016dropout}, rank-1 BNN Gaussian ensemble \cite{RANK-1-BNN}, MIMO \cite{MIMO}, BatchEnsemble \cite{Wen2020BatchEnsemble}, TreeNet \cite{lee2015treenet}, static sparse ensemble of $M$ static sparse networks, Lottery Ticket Rewinding (LTR) ensemble, pruning and fine-tuning ensemble~\cite{han2015PFens}, DST and EDST ensembles~\cite{FREE-TICKETS}, and ensembles of dense DNNs and BNNs. To ensure a fair comparison, all models were trained under consistent hardware, environment, data augmentation, and training schedules. We also imported many of the ensemble model results from \cite{FREE-TICKETS}. Further details on model implementation and learning parameters can be found in Appendix~\ref{App:Expt_details}.

\vspace{1mm}
\noindent \textbf{Metrics.} We quantify predictive performance and robustness focusing on the accuracy (Acc), negative log-likelihood (NLL), and expected calibration error (ECE) on the i.i.d. test data (CIFAR-10 and CIFAR-100) and corrupted test data (CIFAR-10-C and CIFAR-100-C) involving 19 types of corruption (e.g., added blur, compression artifacts, frost effects) \cite{CIFAR-C}. Additional details on the evaluation metrics are given in the Appendix~\ref{App:Expt_details}.

\vspace{1mm}
\noindent \textbf{Results.} The results for CIFAR-10 and CIFAR-100 experiments are presented in Tables~\ref{table:wide_resnet_cifar_10} and \ref{table:wide_resnet_cifar_100}, respectively. We choose the ensemble size $M$ in our {\tt SeBayS} approach same as the ones used in EDST method \cite{FREE-TICKETS} for a fair comparison. We report the results for single training pass models in the upper half and multiple training pass models in the lower half of the Tables~\ref{table:wide_resnet_cifar_10} and \ref{table:wide_resnet_cifar_100}.

We included two {\tt SeBayS} ensembles with different sparsity levels. In the first, we set $S=0.8$, making 80\% of the weights sparse and collecting $M=3$ subnetworks for ensembling. In the second, we increased sparsity to $S=0.9$, allowing us to collect $M=7$ subnetworks within a similar training budget. In Table~\ref{table:wide_resnet_cifar_10}, we show that the {\tt SeBayS} ensemble outperforms other single-pass models in terms of accuracy and robustness on CIFAR-10. Moreover, the {\tt SeBayS} ensemble surpasses both the dense BNN ensemble and the {\tt BayS} ensemble, achieving the lowest expected calibration error (ECE) among all the models for CIFAR-10.

In Table~\ref{table:wide_resnet_cifar_100}, we observe that the {\tt SeBayS} ensemble with $S=0.9$ achieves the lowest ECE on CIFAR-100 and CIFAR-100-C among the single pass models. Meanwhile, the {\tt SeBayS} ensemble with $S=0.8$ achieves the lowest negative log-likelihood (NLL) on CIFAR-100-C among the single pass models. Interestingly, the {\tt BayS} ensemble achieves the lowest ECE on CIFAR-100 and CIFAR-100-C, as well as the lowest NLL on CIFAR-100-C among all the models, while maintaining strong predictive performance. Notably, both the {\tt SeBayS} and {\tt BayS} ensembles outperform the dense BNN ensemble on the ECE metric, with the {\tt BayS} ensemble also yielding a lower NLL compared to the dense BNN ensemble.

Lastly, we have performed out-of-distribution (OoD) detection and adversarial robustness studies which we have summarized in Appendices~\ref{App:Ood_robustness} and \ref{App:adversarial_robustness}.

\section{Ensemble Analysis}
In this section, we demonstrate that the subnetworks converge to distinct local optima and exhibit functional behavior similar to that of independently trained sparse BNNs.
\subsection{Function Space Analysis}
{\it Functional space analysis} examines how neural networks map inputs to outputs within function space.

\noindent \textbf{Quantitative Metrics.} We measure the diversity of the sparse base learners in our {\tt SeBayS} ensemble by quantifying the pairwise similarity of the base learner's predictions on the test data. The average pairwise similarity is given by $\mathcal{D}_d = \mathbb{E}\left[d(\mathcal{P}_1(y|x_1,\cdots,x_N), \mathcal{P}_2(y|x_1,\cdots,x_N)) \right]$ where $d(.,.)$ is a distance metric between the predictive distributions and $\{(\boldx_i,y_i)\}_{i=1,\cdots,N}$ are the test data. We consider two distance metrics: 

\noindent (1) \textit{Disagreement:} the fraction of the predictions on the test data on which the base learners disagree: $d_{\rm dis}(\mathcal{P}_1,\mathcal{P}_2) = \frac{1}{N} \sum_{i=1}^N \indicator(\text{arg max}_{\hat{y_i}} \mathcal{P}_1(\hat{y_i}) \neq \text{arg max}_{\hat{y_i}} \mathcal{P}_2(\hat{y_i}))$. 

\noindent (2) \textit{Kullback-Leibler (KL) divergence}: $d_{\rm KL}(\mathcal{P}_1,\mathcal{P}_2)=\mathbb{E}\left[\log \mathcal{P}_1(y) - \log \mathcal{P}_2 (y) \right]$. 

When two models produce identical predictions for all the test data, then both disagreement and KL divergence are zero.

\begin{table}[htb]
\fontsize{9}{11}\selectfont
\centering
\setlength{\tabcolsep}{2.5pt}
\begin{tabular}{l*{6}{c}}
    \toprule
     & \multicolumn{3}{c@{\extracolsep{2.8pt}}}{CIFAR-10 Experiment} & \multicolumn{3}{c@{\extracolsep{2.8pt}}}{CIFAR-100 Experiment} \\
    \midrule
     {Ensemble} & $d_{\rm dis}$ ($\uparrow$) & $d_{\rm KL}$ ($\uparrow$) & Acc ($\uparrow$) & $d_{\rm dis}$ ($\uparrow$) & $d_{\rm KL}$ ($\uparrow$) & Acc ($\uparrow$) \\
    \midrule
    LTR* & 0.026 & 0.056 & 96.2 & 0.111 & 0.185 & 82.1 \\
    Static* & 0.031 & 0.079 & 96.0 & 0.156 & 0.401 & 82.4 \\
    EDST* & 0.031 & 0.073 & 96.3 & 0.126 & 0.237 & 82.2 \\
    MIMO* & 0.032 & 0.086 & 96.4 & -- & -- & 82.0 \\ 
    PF* & 0.035 & \textbf{0.103} & 96.4 & 0.148 & 0.345 & 83.2 \\
    DST* & 0.035 & 0.095 & 96.3 & 0.166 & 0.411 & \textbf{83.3} \\
    \RE{SeBayS} & \RE{0.035} & \RE{0.082} & \RE{96.2} & \RE{0.148} & \RE{0.280} & \RE{81.5} \\
    \RE{BayS} & \RE{\textbf{0.038}} & \RE{0.102} & \RE{\textbf{96.5}} & \RE{\textbf{0.178}} & \RE{\textbf{0.434}} & \RE{82.0} \\
    \midrule
    DNN* & 0.032 & 0.086 & 96.6 & 0.145 & 0.338 & 82.7 \\
    \RE{BNN} & \RE{0.038} & \RE{0.104} & \RE{96.5} & \RE{0.164} & \RE{0.407} & \RE{82.5} \\
    \bottomrule
\end{tabular}
\caption{Diversity metrics: prediction disagreement $(d_{\rm dis})$ and KL divergence $(d_{\rm KL})$ among sparse ensembles (M = 3, S = 0.8). Results with * are obtained from \cite{FREE-TICKETS}. DNN and BNN represent the dense DNN and BNN ensembles.}
\label{table:diversity_analysis}
\end{table}

\begin{figure*}[htb]
\centering
\begin{subfigure}[b]{0.235\textwidth}
    \centering
    \includegraphics[width=\textwidth]{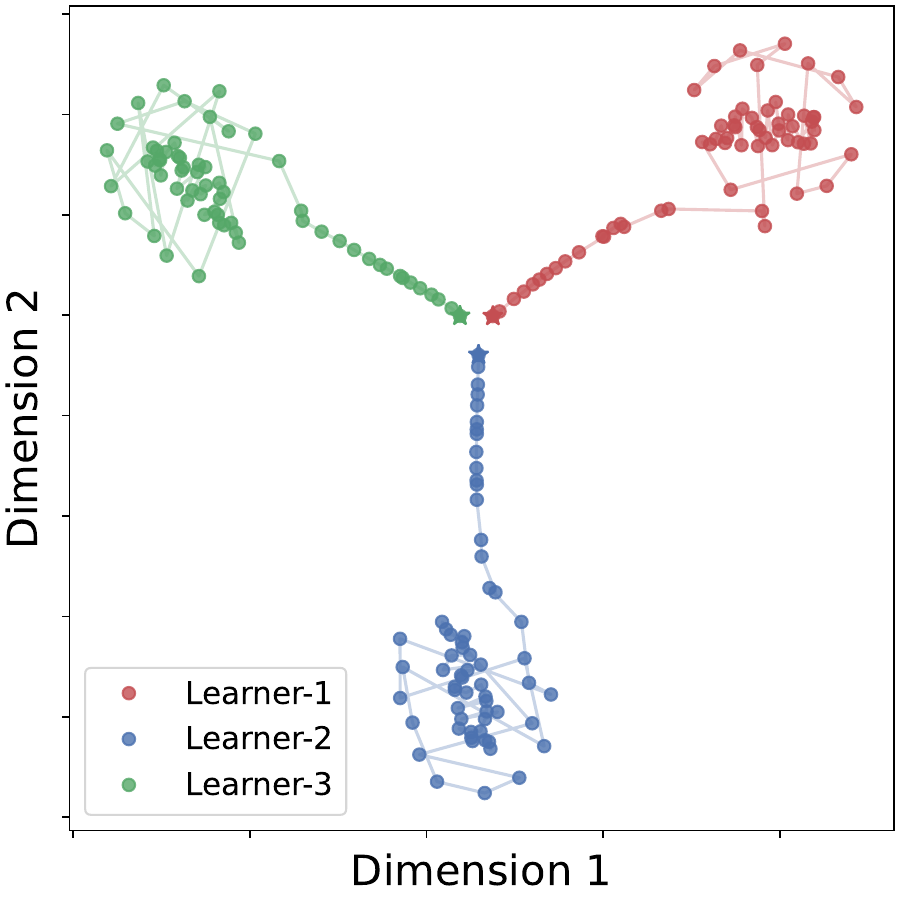} 
    \caption[]%
    {{\small \texttt{BayS} on CIFAR-10}}    
    \label{fig:BayS_C10_TSNE}
\end{subfigure}
\hfill
\begin{subfigure}[b]{0.235\textwidth}  
    \centering 
    \includegraphics[width=\textwidth]{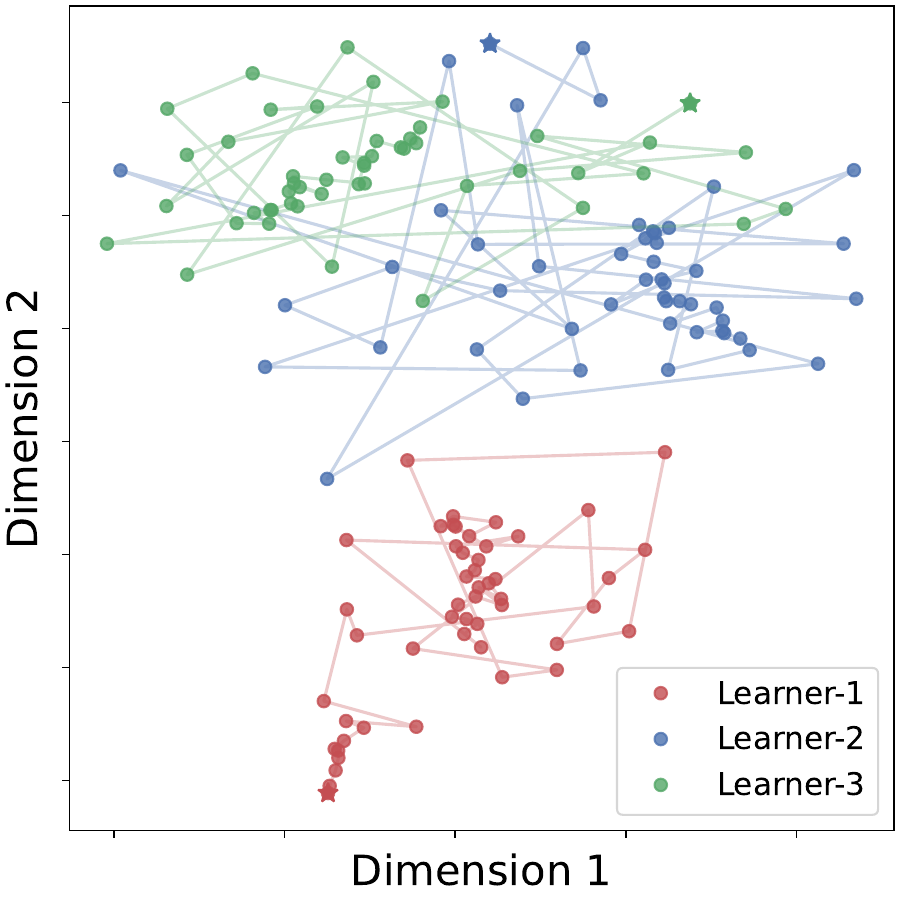} 
    \caption[]%
    {{\small \texttt{SeBayS} on CIFAR-10}}    
    \label{fig:SeBayS_C10_TSNE}
\end{subfigure}
\hfill
\begin{subfigure}[b]{0.235\textwidth}  
    \centering 
    \includegraphics[width=\textwidth]{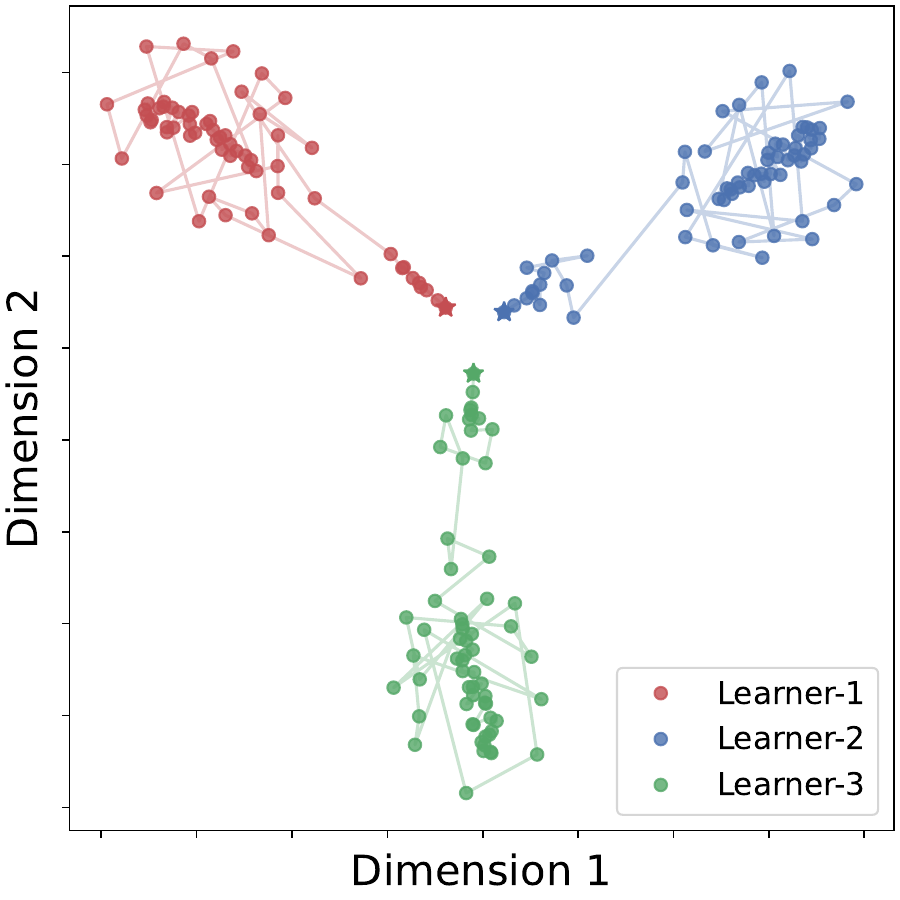} 
    \caption[]%
    {{\small \texttt{BayS} on CIFAR-100}}   
    \label{fig:BayS_C100_TSNE}
\end{subfigure}
\hfill
\begin{subfigure}[b]{0.235\textwidth}  
    \centering 
    \includegraphics[width=\textwidth]{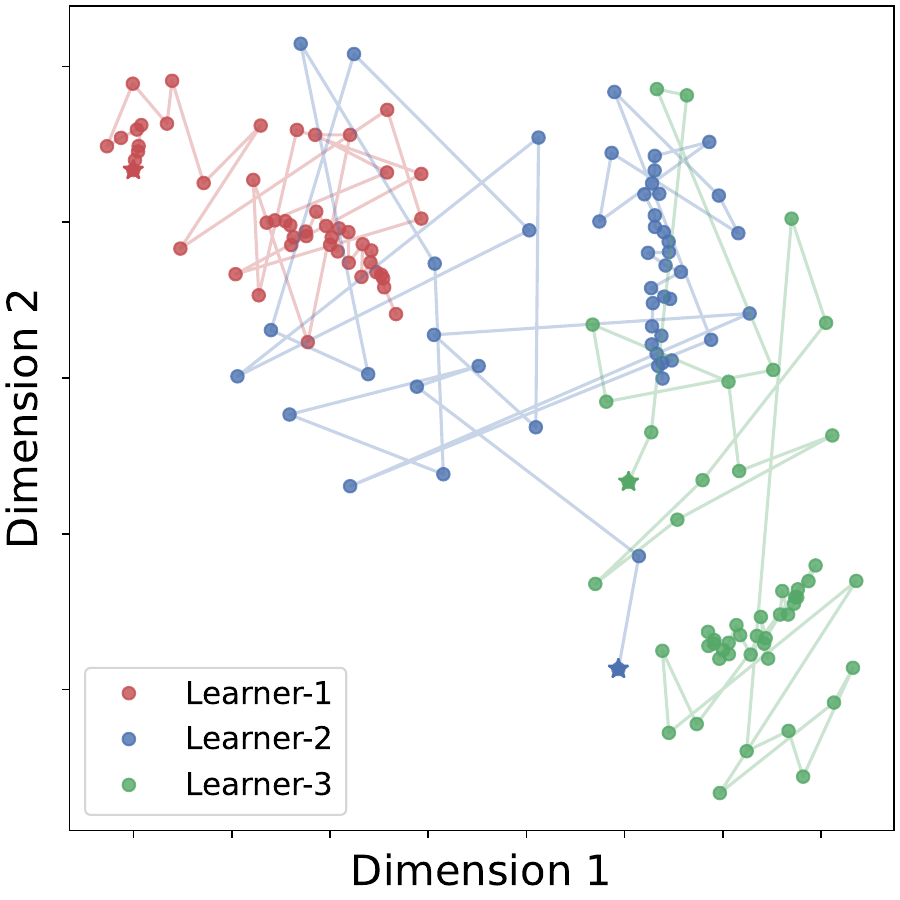} 
    \caption[]%
    {{\small \texttt{SeBayS} on CIFAR-100}}   
    \label{fig:SeBayS_C100_TSNE}
\end{subfigure}
\caption{Training trajectories of base learners obtained by parallel and sequential ensembling of Bayesian subnetworks -- \texttt{BayS} Ensemble and \texttt{SeBayS} Ensemble in Wide ResNet28-10 on CIFAR-10 and CIFAR-100 experiments.}
\label{fig:t-SNE}
\end{figure*}

We report the results of the diversity analysis of the base learners that make up {\tt SeBayS} and {\tt BayS} ensembles ($M=3, S=0.8$) in Table~\ref{table:diversity_analysis} and compare them with the other sparse ensemble methods. We observe that {\tt BayS} ensemble achieves diversity that surpasses all the methods except in CIFAR-10 case it is comparable to dense BNN ensemble and PF ensemble on KL metric. Our single pass {\tt SeBayS} ensemble achieves diversity metrics which surpass the single pass EDST ensemble, multiple pass LTR, and Static ensembles, and are comparable with other single and multiple pass methods. This highlights the importance of dynamic sparsity learning during each exploitation phase to collect diverse set of Bayesian subnetworks in our ensembling approach.

\vspace{1.2mm}
\noindent \textbf{Training Trajectory.} We use t-SNE \cite{TSNE-Paper} to visualize the training trajectories of the sparse base learners generated by our sequential ensembling strategy in the functional space. In our WideResNet28-10 on CIFAR-10/100 experiments, we periodically save checkpoints for each subnetwork following the exploration phase and collect predictions on the test dataset at these checkpoints. After training, we use t-SNE to project these predictions into 2D space. As shown in Figure~\ref{fig:t-SNE}, the local optima achieved by the individual base learners in the {\tt BayS} ensemble are distinctly different, reflecting the high diversity of the ensemble. In contrast, the base learners in the {\tt SeBayS} ensemble converge to relatively closer local optima.


\subsection{Effect of Ensemble size}
In this section, we examine the effect of the ensemble size $M$ in WideResNet28-10 on CIFAR-10 experiment. For {\tt SeBayS} ensemble, we adjust sparsity to maintain consistent training costs across different ensemble sizes. In contrast, the {\tt BayS} ensemble trains $M$ subnetworks in parallel, using the same sparsity level as {\tt SeBayS} ensemble for the corresponding ensemble size. According to the ensembling literature \cite{NN-ensemble-1990,Ovadia-uncertainty-2019}, increasing number of diverse base learners in the ensemble improves predictive performance, although with a diminishing impact. We generate models and aggregate performance with increasing $M$.

\begin{figure}[htb]
\centering
\begin{subfigure}[b]{0.23\textwidth}  
    \centering 
    \includegraphics[width=\textwidth]{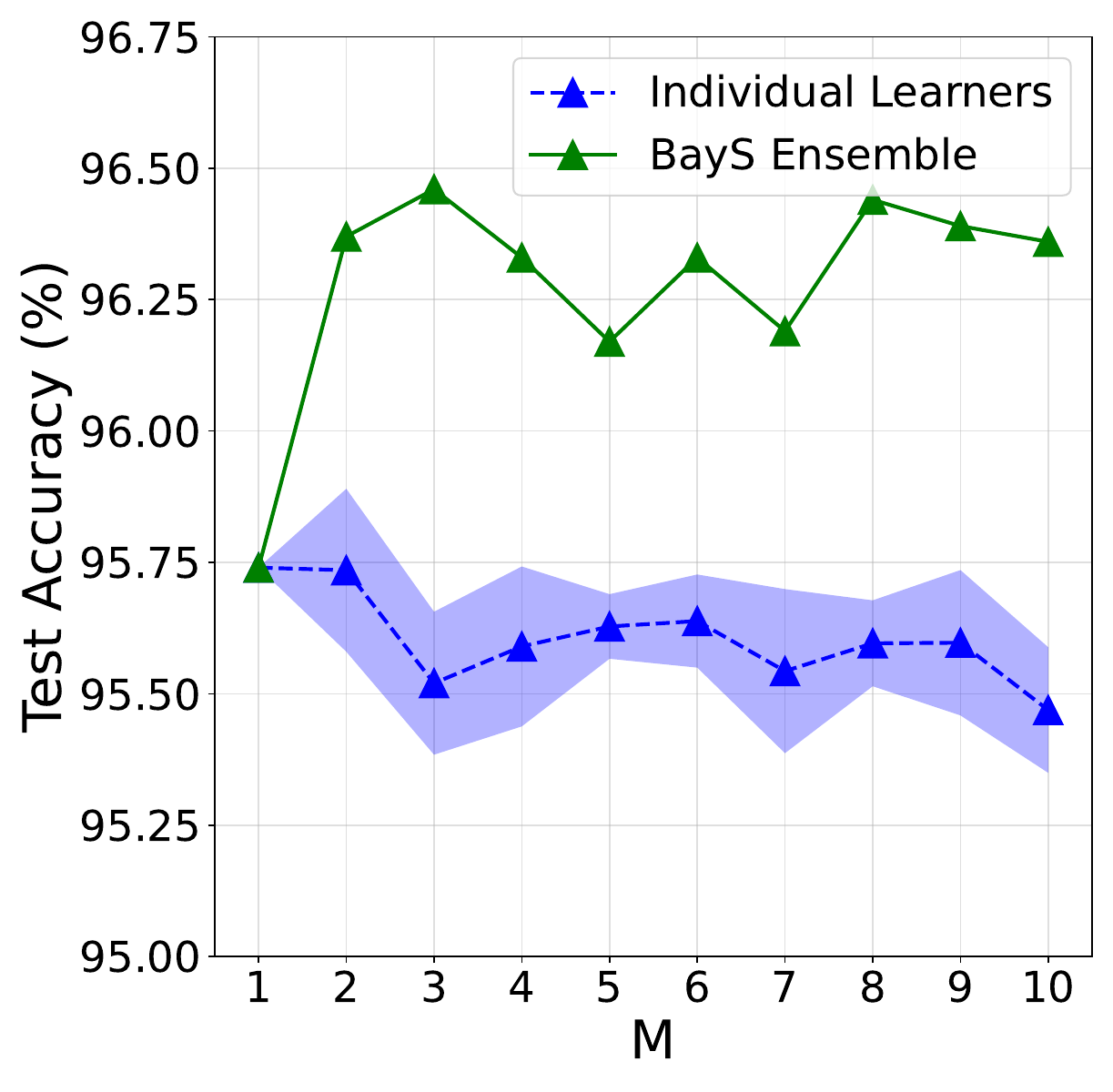} 
    \caption[]%
    {{\small \texttt{BayS} Ensemble}}    
\end{subfigure}
\hfill
\begin{subfigure}[b]{0.23\textwidth}  
    \centering 
    \includegraphics[width=\textwidth]{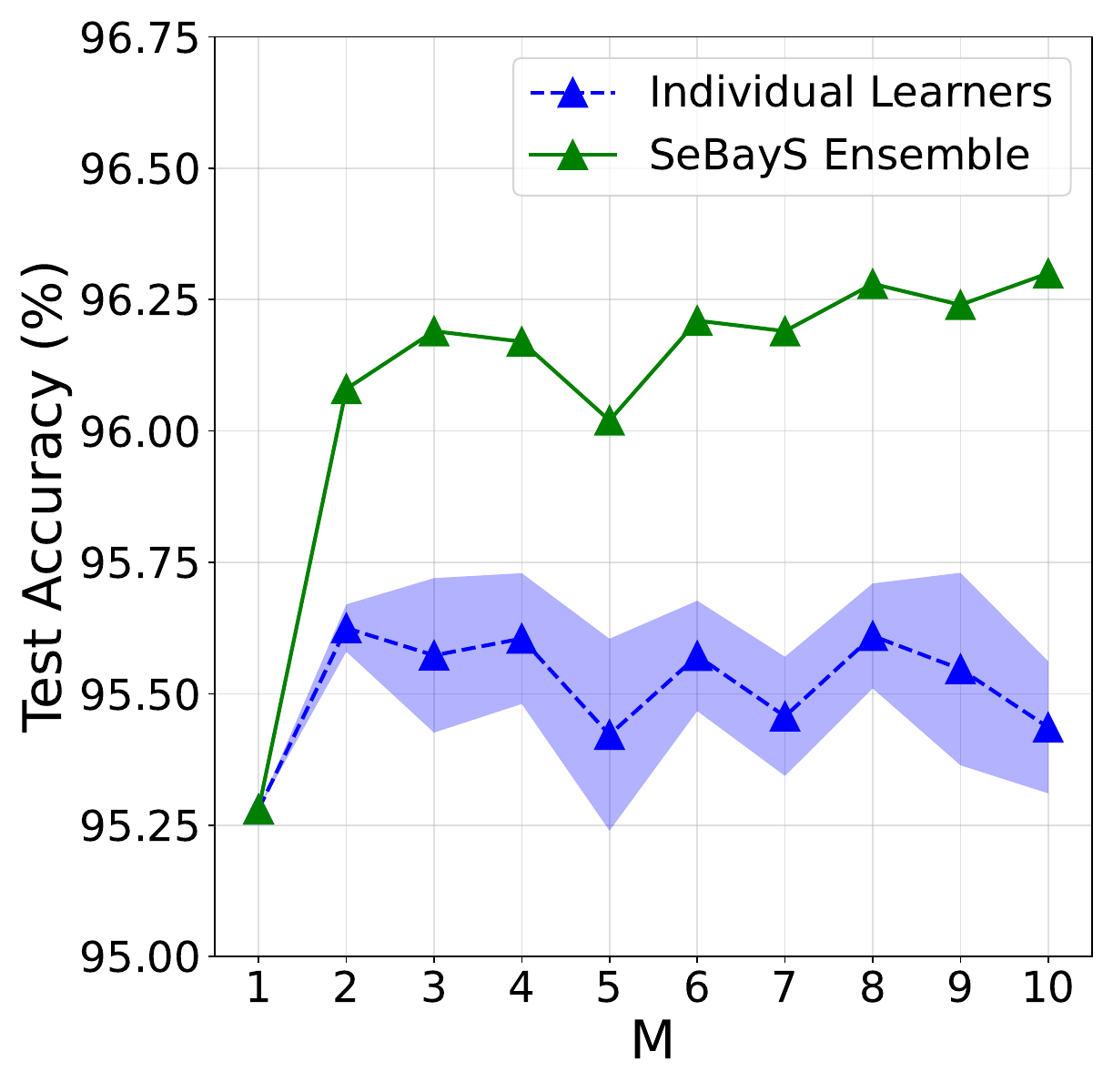} 
    \caption[]%
    {{\small \texttt{SeBayS} Ensemble}}    
\end{subfigure}
\caption{Performance of base learners and their ensembles as ensemble size M varies in CIFAR-10 experiment.}
\label{fig:ensemble-size-effect}
\end{figure} 

In Figure~\ref{fig:ensemble-size-effect}, we plot the performance of the individual learners and their ensembles with varying $M$. For individual learners, we show the mean test accuracy with one standard deviation spread. When $M=1$, the ensemble and individual model refer to a single learner and their performance match. As $M$ grows, we observe that {\tt SeBayS} and {\tt BayS} ensembles maintain their performance while outperforming single dense BNN. The high performance of our {\tt SeBayS} ensemble compared to its individual learners highlights the advantages of our efficient sequential ensembling approach.
\section{Conclusion and Discussion}
\label{sec:conclusion_discussion}

In this work, we propose the \texttt{SeBayS} ensemble, an approach that generates sequential Bayesian neural subnetwork ensembles by combining a novel sequential ensembling technique for BNNs with dynamic sparsity learning. By maintaining fully sparse connectivity throughout training, this method offers a simple yet effective way to enhance predictive performance and model robustness. The highly diverse Bayesian neural subnetworks converge to different optima in the function space, and when combined, they form an ensemble that demonstrates improved performance compared to single dense Bayesian neural network. We have conducted extensive experiments showing that our {\tt SeBayS} method outperforms other ensemble methods in accuracy, uncertainty quantification, out-of-distribution (OoD) detection, and adversarial robustness, all while remaining computationally efficient.

Future work could explore energy-efficient large-scale uncertainty estimation frameworks to further reduce the computational burden. Although dynamic sparsity learning is effective in reducing computational complexity, optimizing sparse structures for current GPU hardware is crucial. Specifically, exploring structured sparsity, such as fine-grained $N:M$ sparsity in ensembling, could be promising.

\appendix

\newpage

\section{Prune-Grow Criterion}  

\label{App:prune-grow}
The Signal-to-Noise Ratio (SNR) is commonly used to identify which weights should be pruned in a Bayesian model by equally considering both the magnitude of the weights and the noise associated with them. It is defined as:
$$ {\rm SNR}_{q_{\phi_i}}(\theta_i) = \frac{|\mu_i|}{\sigma_i} $$
Here, $\mu_i$ and $\sigma_i$ denote the variational mean and standard deviation parameters associated with weight $\theta_i$. Specifically, $|\mu_i|=|\E_{q_{\phi_i}}(\theta_i)|$ and $\sigma_i={\rm Var}_{q_{\phi_i}}(\theta_i)$. However, in Bayesian Neural Networks, weights are sampled from their variational distribution before each forward pass. Therefore, rather than using the absolute value of the average weights/signals scaled by their variations, it is more appropriate to use the average magnitude of signals scaled by their corresponding variation. This approach is preferred because weights with smaller absolute values have a reduced influence on the network's output, on average. In our proposed approach, we instead use ${\rm SNR}$ of $|\theta_i|$, similar to \cite{li2024ssvi}, which is defined as:
$$ {\rm SNR}_{q_{\phi_i}}(|\theta_i|) = \frac{\E_{q_{\phi_i}}|\theta_i|}{\sqrt{{\rm Var}_{q_{\phi_i}}|\theta_i|}} $$
In what follows we drop the weight index $i$ for convenience. 

\vspace{1.2mm}
\noindent First we derive the $\E_{q_{\phi}}|\theta|$ expression.
\begin{align*}
    \E_{q_{\phi}}|\theta| &= \int_{\R} |\theta| \frac{1}{\sqrt{2 \pi \sigma^2}} \exp \left( - \frac{(\theta-\mu)^2}{2\sigma^2} \right ) d\theta \\
    & {\rm let} \enskip z=\frac{x-\mu}{\sigma} \\
    &= \int_{\R} |\mu + \sigma z| \frac{1}{\sqrt{2 \pi}} \exp \left( - \frac{z^2}{2} \right ) dz \\
    & {\rm let} \enskip \psi(z)=\frac{1}{\sqrt{2 \pi}} \exp \left( - \frac{z^2}{2} \right ) \\
    &= \mu \left\{ - \int_{-\infty}^{-\frac{\mu}{\sigma}} \psi(z) dz + \int_{-\frac{\mu}{\sigma}}^{\infty} \psi(z) dz \right \} \\
    & \quad + \sigma \left\{ - \int_{-\infty}^{-\frac{\mu}{\sigma}} z \psi(z) dz + \int_{-\frac{\mu}{\sigma}}^{\infty} z \psi(z) dz \right \} \\
    & {\rm let} \enskip \Phi(z)= \int_{-\infty}^{z} \psi(z) dz \\
    & \quad \left ( \Phi(z) \text{ is the cumulative distribution function of} \right. \\
    & \quad \enskip \left. \text{ standard normal distribution} \right) \\
    &= \mu \left[ -\Phi \left(-\frac{\mu}{\sigma} \right) + \Phi \left(\frac{\mu}{\sigma} \right) \right] + \sigma \sqrt{\frac{2}{\pi}} \exp \left( -\frac{\mu^2}{2\sigma^2} \right ) \\
    & \text{since, } \Phi(-z) = 1 - \Phi(z) \\
    &= \mu \left( 2 \Phi \left(\frac{\mu}{\sigma} \right) - 1 \right) + \frac{2\sigma}{\sqrt{2\pi}} \exp \left( -\frac{\mu^2}{2\sigma^2} \right )
\end{align*}
Next we derive the ${\rm Var}_{q_{\phi}}|\theta|$ expression.
\begin{align*}
    {\rm Var}_{q_{\phi}}|\theta| &= \E_{q_{\phi}}|\theta|^2 - \left(\E_{q_{\phi}}|\theta| \right)^2 \\
    &= \E_{q_{\phi}}\theta^2 - \left(\E_{q_{\phi}}|\theta| \right)^2 \\
    &= \sigma^2 + \mu^2 - \left(\E_{q_{\phi}}|\theta| \right)^2 \\
    &= \sigma^2 + \mu^2 \\
    & \enskip - \left[ \mu \left( 2 \Phi \left(\frac{\mu}{\sigma} \right) - 1 \right) + \frac{2\sigma}{\sqrt{2\pi}} \exp \left( -\frac{\mu^2}{2\sigma^2} \right ) \right]^2
\end{align*}
Finally, ${\rm SNR}_{q_{\phi}}(|\theta|)$ is equal to:
$$ \frac{\mu \left( 2 \Phi \left(\frac{\mu}{\sigma} \right) - 1 \right) + \frac{2\sigma}{\sqrt{2\pi}} \exp \left( -\frac{\mu^2}{2\sigma^2} \right )}{\sqrt{\sigma^2 + \mu^2 - \left[ \mu \left( 2 \Phi \left(\frac{\mu}{\sigma} \right) - 1 \right) + \frac{2\sigma}{\sqrt{2\pi}} \exp \left( -\frac{\mu^2}{2\sigma^2} \right ) \right]^2}} $$
\section{Reproducibility Considerations}
\label{App:Expt_details}
\subsection{Hyperparameters}
\noindent {\bf Hyperparameters for {\tt BayS} and dense BNN ensembles.} For Wide ResNet28-10 on CIFAR-10/-100, we use minibatch size of 128 for both models. We train  variational mean parameters for each member of Dense BNN and {\tt BayS} ensemble for 250 epochs with a learning rate of 0.1 which is decayed by a factor of 0.1 at epochs 125 and 188. The variational variance parameter is trained using a learning rate of 0.01 during exploration phase and follows learning rate schedule of variational mean parameters during exploitation phase.
For the {\tt BayS} ensemble, we take the sparsity $S=0.8$, the update interval $\Delta T=1000$, and the prune-grow rate $p=0.5$, same as DST ensemble of \cite{FREE-TICKETS}. We train these models using SGD algorithm with weight decay $5\times10^{-4}$ and momentum $0.9$. Lastly, we have performed extensive hyperparameter search for the {\tt BayS} and dense BNN ensemble models and summarize those choices in Table~\ref{table:hyperparameters}.

\begin{table*}[htb]
\centering
\begin{tabular}{c|ccc}
\toprule
 & & \multicolumn{2}{c}{Experiments} \\
\cmidrule(lr){3-4} 
Methods & Hyperparameter & CIFAR-10 & CIFAR-100 \\
\midrule
\multirow{4}{*}{\RE{BayS Ensemble (M = 3) (S = 0.8)}} & Prior Variance & $1.0$ & $0.04$ \\
& KL annealing steps & 0 & 0 \\
& Exploration phase epochs & 125 & 125 \\
& Single Exploitation phase epochs & 125 & 125 \\
\midrule
\multirow{4}{*}{\RE{SeBayS Ensemble (M = 3) (S = 0.8)}} & Prior Variance &  $0.04$ & $0.04$ \\
& KL annealing steps & 150 & 150 \\
& Exploration phase epochs & 150 & 150 \\
& Single Exploitation phase epochs & 100 & 100 \\
\midrule
\multirow{4}{*}{\RE{SeBayS Ensemble (M = 7) (S = 0.9)}} & Prior Variance &  $0.04$ & $0.04$\\
& KL annealing steps & 0 & 0 \\
& Exploration phase epochs & 150 & 150 \\
& Single Exploitation phase epochs & 100 & 100 \\
\midrule
\multirow{4}{*}{\RE{Dense BNN Ensemble (M = 3)}} & Prior Variance &  $1.0$ & $1.0$ \\
& KL annealing steps & 150 & 0 \\
& Exploration phase epochs & 125 & 125 \\
& Single Exploitation phase epochs & 125 & 125 \\
\bottomrule
\end{tabular}
\caption{We report the hyperparameters for {\tt SeBayS}, {\tt BayS}, and dense BNN ensembles in both CIFAR experiments.}
\label{table:hyperparameters}
\end{table*}

\noindent {\bf Hyperparameters for {\tt SeBayS} ensemble.} For Wide ResNet28-10 on CIFAR-10/-100, the minibatch size is 128. We train {\tt SeBayS} ensemble with $S=0.8$ for 450 epochs generating $M=3$ subnetworks and $S=0.9$ for 850 epochs generating $M=7$ subnetworks. The exploration phase is run for $t_0=150$ epochs and each exploitation phase is run for $t_{\rm ex}=100$ epochs. During the exploration phase, we take a high learning rate of 0.1 for variational mean parameters while variational variance parameter is trained using a learning rate of 0.01. For each exploitation phase, we use learning rate of 0.01 for first $t_{\rm ex}/2=50$ epochs and 0.001 for remaining $t_{\rm ex}/2=50$ epochs for both variational mean and variance parameters. We choose the update interval $\Delta T=1000$, the prune-grow rate $p=0.5$, and large prune-grow rate $p_L=0.8$, same as EDST ensemble of \cite{FREE-TICKETS}. We train the model using SGD algorithm with weight decay $5\times10^{-4}$ and momentum $0.9$. Finally, we have performed extensive hyperparameter search for the {\tt SeBayS} ensemble models and summarize those choices in Table~\ref{table:hyperparameters}.

\vspace{1.2mm}
\noindent {\bf Hyperparameters for single models.} The single dense and sparse models presented in Tables~1 and 2 which include single dense BNN, SeBayS $(M=1,S=0.9)$, SeBayS $(M=1,S=0.8)$, and BayS $(M=1,S=0.8)$ are the best performing models out of their corresponding ensembles.

\vspace{1.2mm}
\noindent {\bf Hyperparameters for other ensemble models.} The model results marked with $^*$ are taken from \citet{FREE-TICKETS}, and we direct readers to Appendix C of their paper for details on the hyperparameter settings used.  Likewise, the results for the rank-1 BNN ensemble are taken from \citet{RANK-1-BNN}, and we recommend consulting their paper for information on hyperparameter choices.

\subsection{Data Augmentation}
For CIFAR-10 and CIFAR-100 train datasets, we first pad the training images using 4 pixels of value 0 on all borders and then crop the padded image at a random location generating train images of the same size as the original train images. Next, with a probability of 0.5, we horizontally flip a given cropped image. Finally, we normalize the images using ${\rm \texttt{mean}}=(0.4914, 0.4822, 0.4465)$ and ${\rm \texttt{standard deviation}} = (0.2470, 0.2435, 0.2616)$ in CIFAR-10 case. Whereas, we use ${\rm \texttt{mean}}=(0.5071, 0.4865, 0.4409)$ and ${\rm \texttt{standard deviation}} = (0.2673, 0.2564, 0.2762)$ in CIFAR-100 case. Next, we split the train data of size 50000 images into a TRAIN/VALIDATION split of 45000/5000 transformed images. For CIFAR-10/100 test data, we normalize the 10000 test images in each data case using the corresponding \texttt{mean} and \texttt{standard deviation} of their respective training data. 

\subsection{Evaluation Metrics}
We quantify the predictive performance of each method using the accuracy of the test data (Acc). For a measure of robustness or predictive uncertainty, we use negative log-likelihood (NLL) calculated on the test dataset. Expected calibration error (ECE) quantifies how well a model's predicted probabilities align with actual outcomes by partitioning predictions into $K$ equally sized bins and computing a weighted average of the absolute differences between the empirical accuracy and predicted confidence within each bin. Moreover, we adopt \{cAcc, cNLL, cECE\} to denote the corresponding metrics on corrupted test datasets. We also use VALIDATION data to determine the best epoch in each model which is later used for TEST data evaluation.

In {\tt SeBayS}, {\tt BayS}, and dense BNN ensemble models, we use one Monte Carlo sample to generate the network parameters and correspondingly generate a single prediction for each individual base learner. We then calculate the ensemble prediction using a simple average of $M$ predictions generated from $M$ base learners and use this averaged prediction to calculate the evaluation metrics mentioned above for the ensemble models. We also experimented with using a higher number of Monte Carlo (MC) samples, $N_{\rm MC}>1$, and computed ensemble predictions by taking a simple average of $M\times N_{\rm MC}$ predictions from all the base learners. However, this approach did not yield significant improvement over using a single MC sample, so we have not included these results here for brevity.

\subsection{Hardware and Software}
We run all the experiments on a single NVIDIA A100 GPU for {\tt SeBayS}, {\tt BayS}, and dense BNN ensemble models.
\newpage
\section{OOD Experiment Results}
\label{App:Ood_robustness}

\begin{table*}[htb]
\centering
  \begin{tabular}{c|cc|cc}
    \toprule
    & \multicolumn{4}{c}{OoD datasets} \\
    \cmidrule(lr){2-5} 
    & \multicolumn{2}{c|}{CIFAR-10 trained models} & \multicolumn{2}{c}{CIFAR-100 trained models} \\
    \cmidrule(lr){2-3} \cmidrule(lr){4-5}
    Methods & SVHN & CIFAR-100 & SVHN & CIFAR-10  \\
    \midrule
    Static Sparse Model (M = 1) (S = 0.8)* & 0.8896 & 0.8939 & 0.7667 & 0.8004 \\
    Static Sparse Ensemble (M = 3) (S = 0.8)* & 0.9229 & 0.9082 & 0.8165 & 0.8141\\
    DST (M = 1) (S = 0.8)* & 0.9082 & 0.8957 & 0.8284 & 0.8019 \\
    DST Ensemble (M = 3) (S = 0.8)* & 0.9533 & 0.9114 & 0.8207 & 0.8221 \\
    EDST (M = 1) (S = 0.8)* & 0.9461 & 0.8895 & 0.7481 & 0.7941 \\
    EDST Ensemble (M = 3) (S = 0.8)* & 0.9487 & 0.9045 & \textbf{0.8585} & 0.8137 \\
    EDST (M = 1) (S = 0.9)* & 0.9439 & 0.8955 & 0.7990 & 0.7937 \\
    EDST Ensemble (M = 7) (S = 0.9)* & \textbf{0.9658} & 0.9115 & 0.8092 & 0.8148 \\
    \RE{BayS (M = 1) (S = 0.8)} & \RE{0.9410} & \RE{0.8845} & \RE{0.8237} & \RE{0.8033} \\
    \RE{BayS Ensemble (M = 3) (S = 0.8)} & \RE{0.9438} & \RE{0.9060} & \RE{0.8158} & \RE{0.8270} \\
    \RE{SeBayS (M = 1) (S = 0.8)} & \RE{0.9294} & \RE{0.8877} & \RE{0.7580} & \RE{0.7988} \\
    \RE{SeBayS Ensemble (M = 3) (S = 0.8)} & \RE{0.9612} & \RE{0.9066} & \RE{0.7520} & \RE{0.8174} \\
    \RE{SeBayS (M = 1) (S = 0.9)} & \RE{0.9405} & \RE{0.8820} & \RE{0.7801} & \RE{0.8004} \\
    \RE{SeBayS Ensemble (M = 7) (S = 0.9)} & \RE{0.9518} & \RE{\textbf{0.9124}} & \RE{0.7924} & \RE{\textbf{0.8232}} \\
    \midrule
    Single Dense DNN Model* & 0.9655 & 0.8847 & 0.7584 & 0.8045 \\
    \RE{Single Dense BNN Model} & \RE{0.9571} & \RE{0.8873} & \RE{0.7995} & \RE{0.8006} \\
    \RE{Dense BNN Ensemble} & \RE{0.9708} & \RE{0.9082} & \RE{0.8047} & \RE{0.8165} \\
    \bottomrule
  \end{tabular}
  \caption{The out-of-distribution (OoD) detection performance measured by ROC-AUC for Wide ResNet28-10 trained on CIFAR-10 and CIFAR-100. Results with * are obtained from \cite{FREE-TICKETS}.}
  \label{table:WRN_Cifar_OoD_expts}
\end{table*}

In Table~\ref{table:WRN_Cifar_OoD_expts}, we present the ROC-AUC results for out-of-distribution (OoD) detection tasks for the Wide ResNet28-10 trained on CIFAR-10 and CIFAR-100 models. We have reported results for single sparse base learners as well as the corresponding ensemble models. We use SVHN \cite{netzer2011svhn} and CIFAR-100 datasets to evaluate the OoD detection performance of the CIFAR-10 trained models. Similarly, we use SVHN and CIFAR-10 as OoD datasets for the CIFAR-100 trained models. We demonstrate that {\tt SeBayS}, {\tt BayS}, and dense BNN ensemble outperform their respective best performing single base learners. Our proposed {\tt SeBayS} ensemble approach shows clear gains over single dense DNN and BNN models in some cases. Finally, {\tt SeBayS} ensemble model produces best results in cases where CIFAR-10 or 100 are the OoD datasets.

\section{Adversarial Robustness Experiments}
\label{App:adversarial_robustness}
In Table~\ref{table:WRN_Cifar_adv_robust_expts}, we present the adversarial robustness results for the Wide ResNet28-10 trained on CIFAR-10 and CIFAR-100 models. We adopt the Fast Gradient Sign Method (FGSM) \cite{szegedy2013intriguing} with step size of $\frac{8}{255}$ similar to \cite{FREE-TICKETS}. To this end, we generated adversarial examples for each model and report the {min, mean, max} robust accuracy across the generated attacks from different models as mentioned in \cite{strauss2017ensemble}. We demonstrate that the {\tt SeBayS} and {\tt BayS} ensembles outperform static sparse, DST, and EDST ensembles in terms of average robust accuracy. Furthermore, the robustness of the {\tt SeBayS} and {\tt BayS} ensembles is comparable to that of single dense DNN and BNN models, with the {\tt BayS} ensemble outperforming the dense BNN ensemble on average.

\begin{table*}[htb]
\centering
\begin{tabular}{c|cc}
\toprule
 & \multicolumn{2}{c}{Robust Accuracy (\%) \{min/mean/max\}} \\
\cmidrule(lr){2-3} Methods & CIFAR-10 & CIFAR-100 \\
\midrule
Static Sparse Ensemble (M = 1) (S = 0.8)* & 33.68/38.40/41.29 & 11.89/15.21/17.01 \\
Static Sparse Ensemble (M = 3) (S = 0.8)* & 39.28/39.46/39.77 & 16.44/16.81/17.19 \\
DST Ensemble (M = 3) (S = 0.8)* & 37.30/38.49/39.12 & 14.96/15.59/15.90 \\
EDST Ensemble (M = 3) (S = 0.8)* & 39.66/40.35/41.00 & 13.00/13.19/13.56 \\
EDST Ensemble (M = 7) (S = 0.9)* & 35.68/37.74/38.67 & 14.45/14.93/15.56 \\
\RE{BayS Ensemble (M = 3) (S = 0.8)} & \RE{\textbf{43.70/43.94/44.21}} & \RE{\textbf{16.97/17.14/17.37}} \\
\RE{SeBayS Ensemble (M = 3) (S = 0.8)}  & \RE{38.42/40.60/41.95} & \RE{13.27/13.77/14.18} \\
\RE{SeBayS Ensemble (M = 7) (S = 0.9)}  & \RE{38.15/39.46/40.32} & \RE{13.78/14.39/14.91} \\
\midrule
Single Dense DNN Model* & 44.20 & 11.82 \\
\RE{Single Dense BNN Model} & \RE{42.95} & \RE{12.81} \\
\RE{Dense BNN Ensemble} & \RE{47.34/48.96/50.51} & \RE{18.77/19.07/19.50} \\
\bottomrule
\end{tabular}
\caption{The robust accuracy (\%) for Wide ResNet-28-10 trained on CIFAR-10 and CIFAR-100. Results with * are obtained from \cite{FREE-TICKETS}.}
\label{table:WRN_Cifar_adv_robust_expts}
\end{table*}

\end{document}